\documentclass{article}
\usepackage{spconf,amsmath,graphicx}

\usepackage[utf8]{inputenc}

\usepackage{booktabs} % for professional tables
\usepackage{multirow}
\usepackage{dblfloatfix} 
\usepackage{subcaption}
\usepackage{pifont}
\usepackage{mathtools}
\usepackage{wrapfig}

\usepackage{relsize}
\usepackage[shortlabels]{enumitem}

\usepackage{microtype}

\usepackage{soul,color}  % soul for highlight and color for, well... colors...

\usepackage{mathtools}
\usepackage{amsfonts}
\usepackage{stmaryrd}
\usepackage{subcaption}

\usepackage{xspace}
\makeatletter
\DeclareRobustCommand\onedot{\futurelet\@let@token\@onedot}
\def\@onedot{\ifx\@let@token.\else.\null\fi\xspace}

\def\ie{\emph{i.e}\onedot} 
 
\def\etc{\emph{etc}\onedot} 
 
\def\etal{\emph{et al}\onedot}
\@namedef{ver@everyshi.sty}{}
\makeatother

\usepackage{relsize}

\def\f{{\bf f}}

\DeclarePairedDelimiter\norm{\lVert}{\rVert}%

\title{Improving memory banks for unsupervised  learning with large mini-batch, consistency and hard negative mining}
\name{Adrian Bulat, Enrique Sánchez-Lozano, Georgios Tzimiropoulos}
\address{Samsung AI Cambridge, Cambridge, UK}
\begin{document}
%\ninept
%
\maketitle
\begin{abstract}
An important component of unsupervised learning by instance-based discrimination is a memory bank for storing a feature representation for each training sample in the dataset. In this paper, we introduce 3 improvements to the vanilla memory bank-based formulation which brings massive accuracy gains: (a) \textit{Large mini-batch}: we pull multiple augmentations for each sample \textit{within the same batch} and show that this leads to better models and enhanced memory bank updates. (b) \textit{Consistency}: we enforce the logits obtained by different augmentations of the same sample to be close without trying to enforce discrimination with respect to negative samples as proposed by previous approaches. (c) \textit{Hard negative mining}: since instance discrimination is not meaningful for samples that are too visually similar, we devise a novel nearest neighbour approach for improving the memory bank that gradually merges extremely similar data samples that were previously forced to be apart by the instance level classification loss. Overall, our approach greatly improves the vanilla memory-bank based instance discrimination and outperforms all existing methods for both seen and unseen testing categories with cosine similarity.

\end{abstract}
\section{Introduction}
\label{sec:intro}

Supervised learning with Deep Neural Networks has been the de facto approach for feature learning in Computer Vision over the last decade. Recently, there is a surge of interest in learning features in an unsupervised manner. This has the advantage of learning from massive amounts of unlabelled/uncurated data for feature extraction and network pre-training and is envisaged to surpass the standard approach of transfer learning from ImageNet or other large labelled datasets. 

The approach we describe in this paper builds upon the widely-used framework of contrastive learning~\cite{ye2019unsupervised,wu2018unsupervised,he2019momentum,oord2018representation,henaff2019data,tian2019contrastive,bachman2019learning} which utilizes a contrastive loss to maximize the similarity between the representations of two different instances of the same training sample while simultaneously minimizing the similarity with the representations computed from different samples. A key point for contrastive learning is the availability of a large number of negative samples for computing the contrastive loss that are stored in a memory bank. Since the memory bank is updated rarely, this is believed to hamper training stability, hence recent methods, like~\cite{ye2019unsupervised,he2019momentum,chen2020simple} advocate online learning without a memory bank.

For this reason, the method of~\cite{ye2019unsupervised} advocates an online approach by defining the positive pair from two differently augmented versions of the same training sample and considers as negatives all other pairs from the same batch, eliminating the memory bank. As opposed to ~\cite{ye2019unsupervised}, we show how to train a powerful network in an unsupervised manner relying on a memory bank-based training approach.  Momentum Contrast~\cite{he2019momentum} maintains and updates a separate encoder for the negative samples rather than storing a memory bank in a fashion similar to the ``mean teacher''~\cite{tarvainen2017mean}. More recently, SimCLR~\cite{chen2020simple} emphasized the importance of composite augmentations, large batch sizes, bigger models and the use of a nonlinear projection head. They suggested that a large minibatch can replace a memory bank. In contrast, our approach employs a memory bank for contrastive learning.

Our main contribution is to show how to massively improve the vanilla memory bank approach of~\cite{wu2018unsupervised} by introducing minimal changes. We explore 2 key ideas: \textbf{(1)} What is the effect of larger batch sizes on contrastive learning with a memory bank? Concurrent work~\cite{chen2020simple} has advocated the use of a large batch size for online training, i.e. \textit{without a memory bank} as it increases the number of negative pairs. We show that a large batch size is also effective for contrastive learning \textit{with a memory bank} (hence decoupling its positive effect from the number of negative pairs) which identifies a connection with gradient smoothing and improved memory bank updates. Furthermore, we show that if a larger mini-batch is constructed so that a set of $K$ augmentations for each instance are used, additional consistency between the instance augmentations can be enforced to further enhance training. \textbf{(2)} Is contrastive learning effective when instances are too visually similar? Intuitively, instance discrimination is not meaningful for such cases. We show that if these samples are ``merged'' into the memory bank, a much more powerful network can be trained. 

When reproducing the evaluation protocol of~\cite{ye2019unsupervised},  we report improvements over~\cite{wu2018unsupervised} of up to $\sim 9\%$ on CIFAR-10 and of up to $\sim 10\%$ on STL-10. Furthermore, with these improvements, our method surpasses~\cite{ye2019unsupervised} and~\cite{misra2019self} by $\sim 6\%$ on CIFAR-10 and by $\sim 3\%$ on STL-10, setting for these datasets a new state-of-the-art. Overall, we make the following \textbf{3 contributions}:
\begin{enumerate}[noitemsep,topsep=0pt,leftmargin=12pt]
    \item 
    We propose a \textbf{large mini-batch} for memory-bank based contrastive learning by pulling, for each sample, a set of \textbf{$K$ augmentations within the same batch}. We show that this approach leads to stronger networks and improves the memory bank representation. (\textbf{Section~\ref{ssec:method-instance-aggregation}}).
    \item
    By having a set of $K$ augmentations in our disposal, we also propose a simple \textbf{consistency} loss which enforces the logits obtained by different augmentations of the same sample to be close enough. Notably, this is achieved without trying to enforce discrimination with respect to the negative samples as proposed by previous approaches (\textbf{Section~\ref{ssec:method-instance-consistency}}).
    \item
    We observe that instance discrimination is not meaningful for samples that are too visually similar. Hence, we propose a \textbf{hard negative mining} approach for improving the memory bank that gradually merges extremely visually similar data samples that were previously forced to be apart by the instance level classification loss (\textbf{Section~\ref{ssec:method-hard-mining}}).
\end{enumerate}

\section{Method}\label{sec:method}

\begin{figure}[t]
	 \centering
        \includegraphics[trim={5cm 5cm 2cm 4.2cm},clip,width=0.5\textwidth]{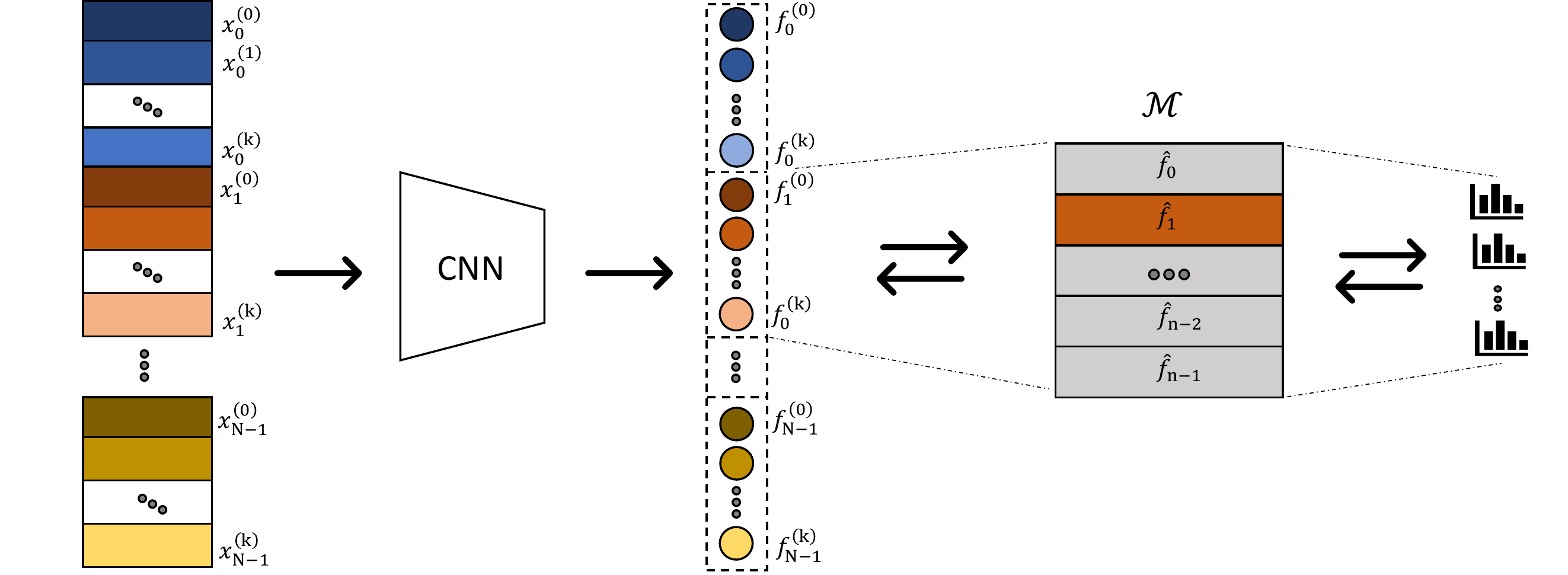}
        \caption{Overall training process. Each instance within the batch is augmented $K$ times and passed as input to the network, producing $N\times K$ embeddings. The final scores are produced by taking the inner product between the feature embedding $\mathbf{f}_i$ and the representations stored in the memory bank $\mathcal{M}.$}
        \label{fig:overall-method-a}
        \vspace{-0.5cm}
\end{figure}

\subsection{Background}\label{ssec:method-preliminaries}

Given a set of $n$ unlabelled images $\mathbf{x}_{1}, \mathbf{x}_{2}, \cdots, \mathbf{x}_{n}$, our goal is to learn a mapping $\Phi(\mathbf{x},\mathbf{\theta})$ from the data to a $d$-dimensional feature embedding $f(\mathbf{x}_{i}) = \Phi(\mathbf{x}_{i},\mathbf{\theta})\in\mathbb{R}^{d}$. Typically $\Phi$ is a neural network and $\theta$ its parameters. Throughout the paper we will simply refer to the feature embedding of the $i$-th sample as $\mathbf{f}_{i}$ and assume that $\norm{\mathbf{f}_i}_{2}=1$. Following~\cite{dosovitskiy2015discriminative,wu2018unsupervised} our pretext task will consist in distinguishing the $i$-th instance from the rest of the samples present in the dataset (\ie each data sample will be treated as a separate class). The training objective is thus formulated as minimizing the negative log-likelihood over all instances of the training set:
\begin{equation}
       \label{eq:loss_single}
    \mathcal{L}_{CE} = \log\prod_{i=1}^{n}P(i|\mathbf{x}_{i}) = \mathlarger{\mathlarger{\sum}_{i=1}^{n}} \log\frac{e^{\mathbf{\hat{f}}_i^T\mathbf{f}_{i}/\tau}}{\sum_{j=1}^{n}e^{\mathbf{\hat{f}}_j^{T}\mathbf{f}_{i}/\tau}}    
\end{equation}
where $\mathbf{\hat{f}}_j$ is a negative sample coming from within the batch~\cite{ye2019unsupervised} or from a memory bank~\cite{wu2018unsupervised}, and $\tau$ is a temperature parameter that controls the concentration of the parameters~\cite{wu2018unsupervised}. 

\begin{figure}[h]
\centering
    \begin{subfigure}[t]{0.2\textwidth}
    \centering
    \includegraphics[trim={4cm 4cm 6cm 3cm},clip,width=1\textwidth]{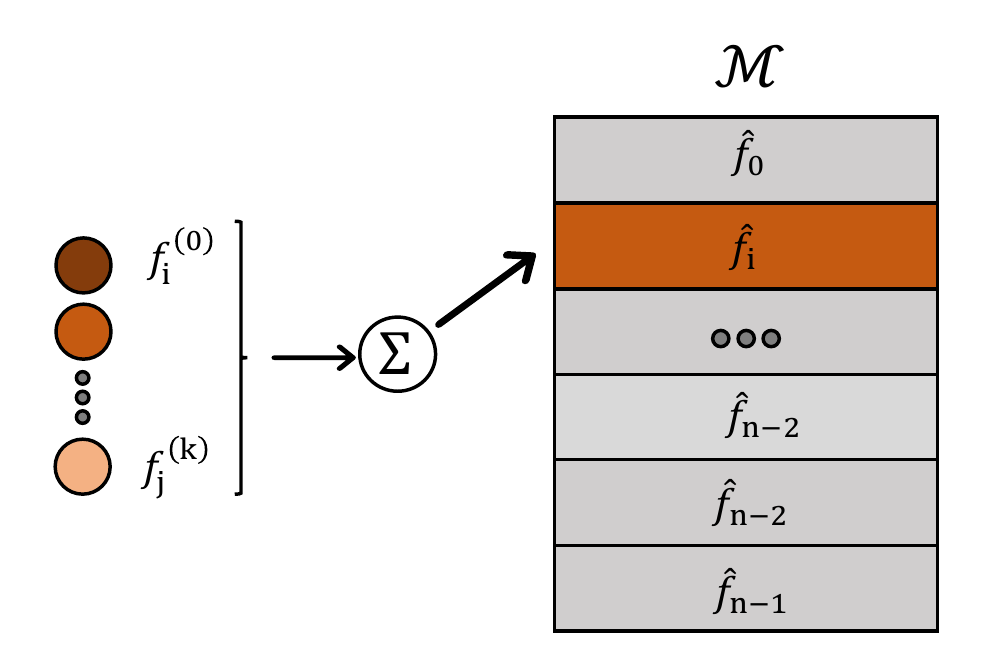}
        \caption{The proposed memory bank update rule shown for a given instance $i$.}
        \label{fig:overall-method-mem-update}
    \end{subfigure}
    ~
    \begin{subfigure}[t]{0.24\textwidth}
        \centering
        \includegraphics[trim={4cm 3.3cm 4cm 3.5cm},clip,width=1\textwidth]{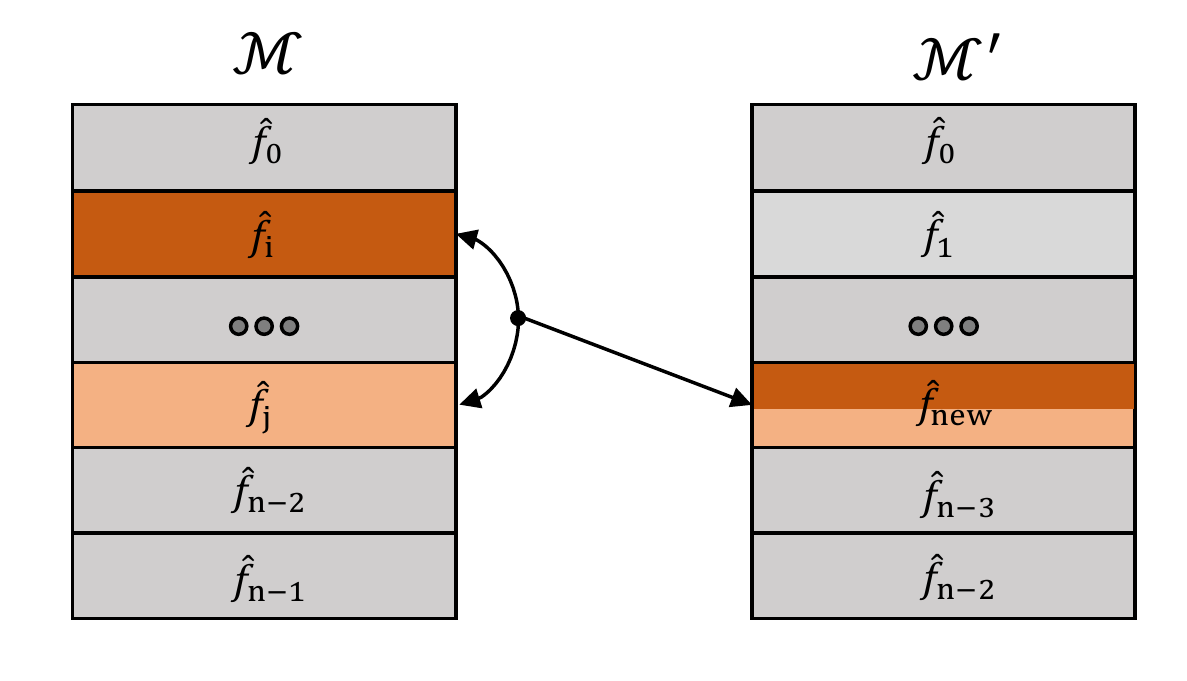}
        \caption{Offline hard mining strategy. Samples with large cosine similarity are merged together.}
        \label{fig:overall-method-d}
    \end{subfigure}
    \caption{Proposed memory bank update mechanisms.}
    \vspace{-0.5cm}
\end{figure}

\subsection{Large mini-batch with multiple augmentations}\label{ssec:method-instance-aggregation}
In contrastive learning, a large mini-batch can be motivated for the case of online learning (no memory bank is used) for increasing the number of negative samples. However, for the case of contrastive learning with a memory bank, the number of negative samples is fixed and independent of the batch size.

\begin{table}
\resizebox{.275\textwidth}{!}{
\parbox{.6\linewidth}{
\centering
\begin{tabular}{c|ccc}
\hline
Expansion & \multicolumn{3}{c}{K}\\
\cline{2-4}
method & 1 & 2 & 4\\
\hline
Standard & \multirow{2}{*}{80.6} & 84.7 & 86.6\\
Multi-augm. &  & \textbf{85.0} & \textbf{87.1} \\
\hline
\end{tabular}
\caption{Top-1 (\%) accuracy on CIFAR-10 for different ways of increasing the batch size.}
\label{table:cifar10-impact-k}
}
}
\hfill
\resizebox{.19\textwidth}{!}{
\parbox{.45\linewidth}{
\centering
\begin{tabular}{l|cc}
\hline
Num.  & \multicolumn{2}{c}{K}\\
\cline{2-3}
feat. & 2 & 4\\
\hline
1 & 83.9 & \textbf{85.0}\\
all & 85.3 & \textbf{87.1} \\
\hline
\end{tabular}
\caption{Top-1 (\%) accuracy on CIFAR-10 vs. number of features used for updating the memory bank.}
\label{table:mem-bank-updates}
}
}
\vspace{-0.5cm}
\end{table}

We make the observation that for the memory-bank case a large mini-batch is useful because it results in more frequent updates for a given feature $\mathbf{\hat{f}}_i$ inside the memory bank. For example if the batch-size is doubled then  ${\mathbf{\hat{f}}}_i$ will be updated twice more frequently. As already mentioned in \cite{wu2018unsupervised}, a memory-bank approach comes at the cost of a large oscillation during training due to inconsistencies caused by updating the feature representations for different samples at very different time instances. Hence, more frequent updates of the memory bank -- offered by a larger mini-batch -- can help stabilize training.
\noindent We consider increasing the batch size by an expansion factor of $K$. There are two ways to achieve this. The \textit{standard} way is to just increase the number of samples at each iteration. All samples, in this case, are different to each other. Table~\ref{table:cifar10-impact-k}
 shows the results obtained by training a network with contrastive learning for $K=1,2,4$ on CIFAR-10 using the kNN evaluation protocol. Clearly, a large batch-size results in much higher accuracy showcasing its benefit in contrastive learning. 
 
The second way to increase the batch size we explore in this work is by using \textit{multiple} -- in particular $K$ -- augmentations per sample \textit{within the same batch}. Specifically, for every input sample $\mathbf{x}_i$ from the batch we propose to construct a series of $K$ perturbed copies $\mathbf{x}_{i}^{(0)},\mathbf{x}_i^{(1)},\cdots,\mathbf{x}_i^{(k)},\cdots,\mathbf{x}_i^{(K-1)}$  using a randomly composed set of augmentations $\mathcal{T}_k$. As such, the loss from one batch $\mathcal{B}$ (with size $|\mathcal{B}|$) becomes:
\begin{equation}
    \label{eq:N-way-softmax-non-parametric}
    \mathcal{L}_{CE}  = \mathlarger{\mathlarger{\sum}_{i=1}^{|\mathcal{B}|}}  \mathlarger{\mathlarger{\sum}_{k=1}^{K}} \log\frac{e^{\mathbf{\hat{f}}_i^T\mathbf{f}_{i}^{(k)}/\tau}}{\sum_{j=1}^{n}e^{\mathbf{\hat{f}}_j^{T}\mathbf{f}_{i}^{(k)}/\tau}},    
\end{equation}
\noindent where $\mathbf{x}_i^{(k)}=\mathcal{T}_{k}(\mathbf{x}_i)$ is the $k$-th augmented copy of image $\mathbf{x}_i$ transformed using a randomly selected set of chained augmentation operators $\mathcal{T}$ (\ie flipping, color jittering \etc) and $f_i^{(k)}$ the corresponding embedding produced by passing the sample $\mathbf{x}_i^{(k)}$ through the network. This is illustrated in Fig.~\ref{fig:overall-method-a} where different shades of the same color represent different augmentations for the same instance. This second way is primarily motivated by being able to enforce the consistency loss described in the next section. The results, shown in Table~\ref{table:cifar10-impact-k}, confirm that by applying the proposed way even higher accuracies can be achieved.

We note that by increasing the batch size in the proposed way (\textit{i.e.} using multiple augmentations) by $K$, the feature $\hat{\f}_i$ is actually updated after the same number of iterations, regardless the value of $K$, which corresponds to the same number of iterations than that of not increasing the batch size. To overcome this issue, we propose the feature $\mathbf{\hat{f}}_i$ to be updated by aggregating the features produced by the $K$ augmented versions of $\mathbf{x}_i$:  $\mathbf{\hat{f}}_i=m\mathbf{\hat{f}}_i+(1-m)\sum_{k=1}^{K}\frac{1}{K}\textbf{f}_i^{(k)}$ (see Fig.~\ref{fig:overall-method-mem-update}). 

The latter observation allows us to further study where the accuracy improvement in Table~\ref{table:cifar10-impact-k} comes from. To this end, we further study the case of using $K$ augmentations to calculate the loss of  Eq. (\ref{eq:N-way-softmax-non-parametric}) but updating the memory bank \textit{only once} (equivalent to using $K=1$). The results for this case for $K=2,4$ are shown in Table~\ref{table:cifar10-impact-k}. Interestingly, we  observe a significant accuracy improvement over the baseline (no augmentation). Since the memory bank is updated in the same way as for the case $K=1$, we conjecture that this accuracy improvement is coming from the \textit{smoothed gradients} due to the use of the large batch size. When measured, the (average) cosine distance between the memory bank representations at adjacent epochs becomes smaller as $K$ increases. Overall, we conclude that a large batch size helps improving both network training and updating the memory bank.

\subsection{Instance consistency}\label{ssec:method-instance-consistency}

\begin{table}
\resizebox{.275\textwidth}{!}{
\parbox{.6\linewidth}{
\centering
\begin{tabular}{lcc}
\hline\noalign{\smallskip}
None & $\ell_2$ & KL (Eq.~\ref{eq:kl})\\
\noalign{\smallskip}
\hline
\noalign{\smallskip}
85.0 & 85.6 & \textbf{86.0} \\
\hline
\end{tabular}
\caption{Top-1 (\%) accuracy on CIFAR-10 obtained using kNN for different methods of consistency regularization for $K=2$ augmentations.}
\label{table:cifar10-knn-consistency}
}
}
\hfill
\resizebox{.19\textwidth}{!}{
\parbox{.45\linewidth}{
\centering
\begin{tabular}{l|c|c}
\hline
\multirow{2}{*}{Training} & \multicolumn{2}{c}{Stage}\\
\cline{2-3}
 & 1 & 2\\
\hline
scratch & 86.5 & 86.7\\
resume & 88.4 & \textbf{89.5} \\
\hline
\end{tabular}
\caption{Top-1 (\%) accuracy on CIFAR-10 using kNN for different stages and training strategies.}
\label{table:mining-strategy}
}
}
\vspace{-0.5cm}
\end{table}

With the introduction of multiple instantiations $\mathbf{x}_i^{(0)}, \mathbf{x}_i^{(1)}, \dots, \mathbf{x}_i^{(k)}$ of the same sample within the batch in the previous subsection, generated by applying a different set of randomly selected transformations $\mathcal{T}_k$, herein we propose to explicitly enforce a consistent representation between the augmented representations of the same image. A similar idea has been explored for the case of semi-supervised learning~\cite{sajjadi2016regularization, miyato2018virtual,berthelot2019remixmatch}, however, to our knowledge, in the context of contrastive learning, this has not explored before. Notably, this consistency is enforced without trying to enforce discrimination with respect to the negative samples as proposed by recent contrastive approaches~\cite{ye2019unsupervised,wu2018unsupervised,he2019momentum,oord2018representation,henaff2019data,tian2019contrastive,bachman2019learning}. More specifically, given a set of logits produced by each of the $K$ augmented copies of $i$, we define our consistency loss as follows:
\begin{equation}
    \label{eq:kl}
    \mathcal{L}_{cons} = \sum_{k=1}^{K}\sum_{j\ne k}\text{KL} \left(P(i|\mathbf{x}_i^{(k)})||P(i|\mathbf{x}_i^{(j)})\right)
\end{equation}

\setlength{\columnsep}{10pt}
\begin{wrapfigure}[12]{r}{0.25\textwidth}
\vspace{-0.4cm}
\centering
    \includegraphics[trim={7cm 5cm 6.5cm 3.5cm},clip,width=0.2\textwidth]{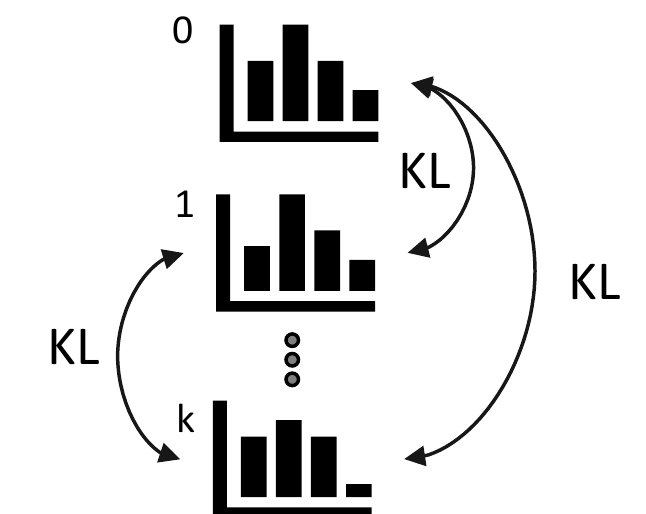}
    \caption{KL consistency loss applied between the logits produced by the $K$ augmented samples of a given sample $i$.}
    \label{fig:overall-method-c}
\end{wrapfigure}
Note that the proposed loss term performs a dense corresponding matching (\ie every possible pair formed using the $K$ augmented samples is considered). This is illustrated in Fig.~\ref{fig:overall-method-c}. For completeness, we also evaluated an $\ell_2$ loss for enforcing consistency. As the results from Table~\ref{table:cifar10-knn-consistency} show, the proposed consistency loss offers noticeable improvements over the \textit{vanilla} training process and the $\ell_2$ form of regularization directly on the feature embeddings.

\subsection{Hard negative  mining}\label{ssec:method-hard-mining}
Unsupervised learning with instance discrimination assumes that a sample within the dataset forms a unique class. An obvious limitation of this approach is that near-identical or very similar samples are artificially forced to be apart in the embedding space. To alleviate this, we propose an offline kNN-based strategy that merges similar instances into a single class. 
As opposed to the deep clustering approach from~\cite{caron2018deep}, we do not seek to construct large clusters in an online manner via K-means nor replacing the instance-level discrimination task; instead, during an offline grouping stage, for each memory bank feature representation, we compute its nearest neighbours, and then group the ones located in its immediate $\sigma$ vicinity (see  Fig.~\ref{fig:overall-method-d}). This process is reminiscent of hard negative mining with the difference being that after the hard negative samples are identified  they are treated as positives. Once the selected instances are merged together they will have a common representation and share the same location inside the memory bank. Similarly, during training, for the grouped instances instead of using $K$ augmentations of the same image, we uniformly sample and  augment images located within the same group. By using a small $\sigma$, the large majority of the samples after grouping stage remain ungrouped (only 5-10\% of  samples are grouped). As such the effect of the proposed approach is to remove very similar samples from being forced to produce different features. Our proposed conservative hard mining strategy is run in an offline manner near the end of the training, each time grouping the most similar samples by means of measuring their cosine distance. Firstly, we notice that the gains flatten out after the algorithm is run for 2 times (\ie denoted as \textbf{stages} in the tables). Secondly, while the method offers improvements even when the model is retrained from scratch using the computed assignments, we find the gains are significantly larger if we continue training from the current checkpoint. Table~\ref{table:mining-strategy} summarizes results showcasing the large impact of our hard negative mining approach.

\section{Experiments}\label{sec:experiments}
We report results for two popular settings: on \textit{seen testing categories} (testing and training is performed on images that contain mutual categories) and \textit{unseen testing categories} (training and testing categories are disjoint). All methods were implemented using PyTorch~\cite{paszke2017automatic}.

\noindent\textbf{Seen Testing Categories.} Following~\cite{wu2018unsupervised,ye2019unsupervised} the experiments are performed on the CIFAR-10~\cite{krizhevsky2009learning} and STL-10~\cite{coates2011analysis} datasets under the same settings. In particular, we use a ResNet18~\cite{he2016deep} as a feature extractor setting the output embedding size to 128. As per~\cite{ye2019unsupervised}, the network is trained for 300 epochs using a starting learning rate of $0.03$, which is then dropped by $0.1$ at epochs 80, 140 and 200. The network is optimized using SGD with momentum ($=0.9$) and a weight decay of $5e-4$. During training each input sample is randomly augmented using a combination of the following transformations: Random resize and crop, random grayscale, random mirroring and color jittering. The temperature $\tau$ is set to $0.1$, the memory bank momentum to $0.5$ and the consistency regularization factor to $\beta = 10^5$. Following~\cite{wu2018unsupervised}, we adhere to the \textit{linear} and \textit{kNN} protocols. As Tables~\ref{table:cifar10-knn} and~\ref{table:stl10} show, our method surpasses other methods, including our direct baseline, the method  of~\cite{wu2018unsupervised}, by a significant margin.

\begin{table}
\centering

\caption{Top-1 (\%) acc. on CIFAR-10 obtained using kNN.}
\label{table:cifar10-knn}
\begin{tabular}{ll}
\hline\noalign{\smallskip}
Method & kNN\\
\noalign{\smallskip}
\hline
\noalign{\smallskip}
Random CNN & 32.1 \\
%DeepCluster (10)~\cite{caron2018deep} & 44.4 \\
DeepCluster (1000)~\cite{caron2018deep} & 67.6 \\
Exemplar~\cite{dosovitskiy2015discriminative} & 74.5 \\
NPSoftmax~\cite{wu2018unsupervised} & 80.8 \\
NCE~\cite{wu2018unsupervised} & 80.4 \\
Triplet~\cite{ye2019unsupervised} & 57.5 \\
Triplet (Hard)~\cite{ye2019unsupervised} & 78.4 \\
Invariant Instance~\cite{ye2019unsupervised} & 83.6 \\
\textbf{Ours} & \textbf{89.5} \\
\hline
\end{tabular}
\vspace{-0.4cm}
\end{table}

\noindent\textbf{Unseen Testing Categories.} Following Song~\etal~\cite{oh2016deep}, we report results by training a ResNet-18 model on unseen categories on the Standford Online Product~\cite{oh2016deep} dataset. The images corresponding to the first half of categories are used for training, in an unsupervised manner, without using their labels, while the testing is done on images belonging to unseen categories. We closely align our setting and training details with~\cite{ye2019unsupervised,movshovitz2017no}: we report results in terms of the clustering quality and NN retrieval performance. We denote with $R@k$ the probability of any correct matching to occur in the top-k retrieved~\cite{oh2016deep}. NMI, the second reported metric, measures the quality of the clustering. As Table~\ref{table:product-pretrain} shows, our method improves in terms of R@1 on top of the state-of-the-art by almost 4\% and on top of our baseline from~\cite{wu2018unsupervised} by 9\%.

\begin{table}
\centering
\caption{Top-1 (\%) acc. on STL-10 using a linear and kNN classifier.}
\label{table:stl10}
\begin{tabular}{llll}
\hline\noalign{\smallskip}
Method & \# img. & Linear & kNN\\
\noalign{\smallskip}
\hline
\noalign{\smallskip}
Random CNN & None & - & 22.4 \\
\hline
HMP*~\cite{bo2013unsupervised} & 105K & 64.5 & - \\
Satck*~\cite{zhao2015stacked} & 105K & 74.3  &- \\
Exemplar*~\cite{dosovitskiy2015discriminative} & 105K & 75.4 &  - \\
Invariant~\cite{ye2019unsupervised} & 105K & 77.9 & 81.6 \\
\hline
NPSoftmax~\cite{wu2018unsupervised} & 5K & 62.3 & 66.8 \\
NCE~\cite{wu2018unsupervised} & 5K & 61.9 & 66.3 \\
DeepCluster (100)~\cite{caron2018deep} & 5K & 56.6 & 61.2 \\
Invariant~\cite{ye2019unsupervised} & 5K & 69.5 & 74.1 \\
\hline
Ours & 5K & 71.9 & 77.6 \\
\textbf{Ours} & 105K & 80.0 & \textbf{84.7} \\
\hline
\end{tabular}
\vspace{-0.2cm}
\end{table}

\begin{table}
\centering
\caption{Results (\%) on Product dataset.}
\label{table:product-pretrain}
\begin{tabular}{lllll}
\hline\noalign{\smallskip}
Method & R@1 & R@10 & R@100 & NMI\\
\noalign{\smallskip}
\hline
\noalign{\smallskip}
Exemplar~\cite{dosovitskiy2015discriminative} & 31.5 & 46.7 & 64.2 & 82.9 \\
NCE~\cite{wu2018unsupervised} & 34.4 & 49.0 & 65.2 & 84.1 \\
MOM~\cite{iscen2018mining} & 16.3 & 27.6 & 44.5 & 80.6 \\
Invariant Instance~\cite{ye2019unsupervised} & 39.7 & 54.9 & 71.0 & 84.7 \\
\textbf{Ours }& \textbf{43.6} & \textbf{57.5} & \textbf{71.8} & \textbf{85.3} \\
\hline
\end{tabular}
\vspace{-0.4cm}
\end{table}

\vspace{-0.2cm}
\section{Conclusion}

We described three simple yet powerful ways to improve unsupervised contrastive learning with a memory bank. Firstly, we proposed a large mini-batch with multiple instance augmentations for providing smoother gradients for improving network training and increasing the quality of the features stored in the memory bank. Secondly, we introduced a simple, yet effective, intra-instance consistency loss that encourages the distribution of each augmented sample to match that of the remaining augmentations. Finally, we presented our very hard mining strategy that attempts to overcome one of the problems of unsupervised instance discrimination: that of trying to push apart near-identical images. We exhaustively evaluated the proposed improvements reporting large accuracy improvements. % over the state-of-the-art.

\clearpage

% References should be produced using the bibtex program from suitable
% BiBTeX files (here: strings, refs, manuals). The IEEEbib.bst bibliography
% style file from IEEE produces unsorted bibliography list.
% -------------------------------------------------------------------------
\bibliographystyle{IEEEbib}
\bibliography{refs}

\end{document}